\documentclass[11pt]{article}

\usepackage[preprint]{acl}

\usepackage{times}
\usepackage{latexsym}

\usepackage[T1]{fontenc}

\usepackage[utf8]{inputenc}

\usepackage{microtype}

\usepackage{inconsolata}

\usepackage{graphicx}

\usepackage{amsmath}
\usepackage{multirow}
\usepackage[table,xcdraw]{xcolor}
\usepackage{booktabs}
\usepackage{algorithm}
\usepackage{algorithmicx}
\usepackage{algpseudocode}
\floatname{algorithm}{Algorithm} 
\usepackage{tabularx}  
\usepackage{array}     

\usepackage{booktabs, multirow, tabularx, siunitx}
\usepackage{enumitem}
\usepackage{makecell}

\title{OptiSet: Unified Optimizing Set Selection and Ranking for Retrieval-Augmented Generation}

\author{
 \textbf{Yi Jiang},
 \textbf{Sendong Zhao\thanks{Corresponding author.}},
 \textbf{Jianbo Li},
 \\
 \textbf{Bairui Hu},
 \textbf{Yanrui Du},
 \textbf{Haochun Wang},
 \textbf{Bing Qin}
\\
\\
 Harbin Institute of Technology, China
 \\
  \texttt{
   \{yjiang,sdzhao,jianboli,brhu,yrdu,hcwang,qinb\}@ir.hit.edu.cn
 }
}

\begin{document}
\maketitle

\begin{abstract}
Retrieval-Augmented Generation (RAG) improves generation quality by incorporating evidence retrieved from large external corpora. However, most existing methods rely on statically selecting top-k passages based on individual relevance, which fails to exploit combinatorial gains among passages and often introduces substantial redundancy. 
To address this limitation, we propose \textbf{OptiSet}, a set-centric framework that unifies set selection and set-level ranking for RAG. OptiSet adopts an ``Expand-then-Refine'' paradigm: it first expands a query into multiple perspectives to enable a diverse candidate pool and then refines the candidate pool via re-selection to form a compact evidence set. 
We then devise a self-synthesis strategy without strong LLM supervision to derive preference labels from the set conditional utility changes of the generator, thereby identifying complementary and redundant evidence.
Finally, we introduce a set-list wise training strategy that jointly optimizes set selection and set-level ranking, enabling the model to favor compact, high-gain evidence sets. 
Extensive experiments demonstrate that OptiSet improves performance on complex combinatorial problems and makes generation more efficient. 
The source code is publicly available\footnote{https://github.com/liunian-Jay/OptiSet.git}.
\end{abstract}

\section{Introduction}

Retrieval-Augmented Generation (RAG) addresses the limitations of fixed knowledge in Large Language Models (LLMs) by integrating an external retrieval component that fetches relevant information from a large corpus~\citep{lewis2020retrieval,zhao2024retrieval1,fan2024survey,li2025matching}. This approach helps overcome challenges like difficulty in updating knowledge and the generation of hallucinations~\citep{bechard2024reducing,gade2025s}, leading to significant improvements in various NLP applications.

\begin{figure}[!t]
    \centering
    \includegraphics[width=0.925\linewidth]{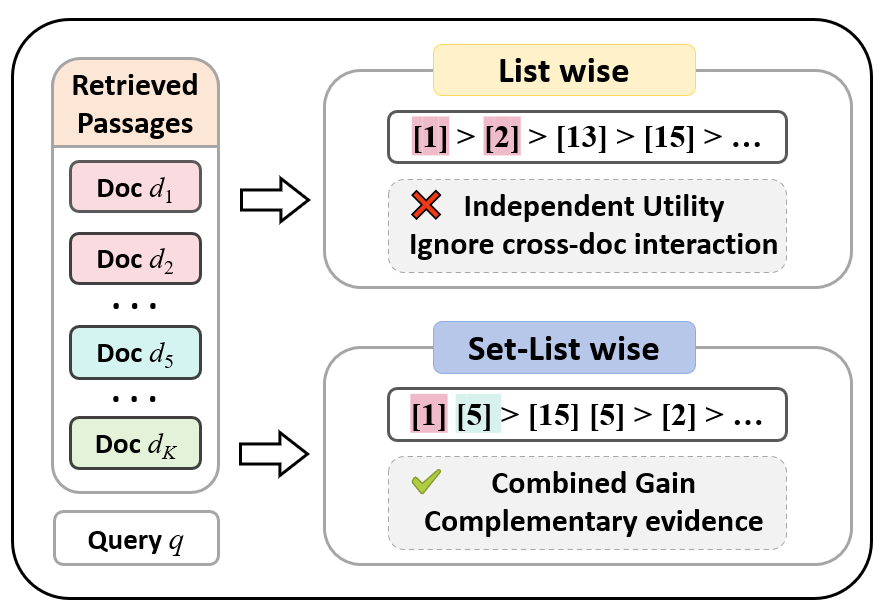}
    \caption{Illustration of the difference in objectives: Set-list wise modeling unifies the modeling of set selection to capture combinatorial gains and reduce redundancy.}
    \label{fig:motivation}
\end{figure}

Many studies have explored the integration of retrieval and generation. For instance, denoising~\citep{wei2025instructrag} and compression~\citep{xu2024recomp} methods aim to reduce redundant information but may compromise information traceability, which is crucial in practical applications. Rerankers~\citep{glass2022re2g}, by re-evaluating passage relevance and selecting the most relevant top-k passages, can seamlessly integrate with and complement various other strategies, and have remained a mainstream research direction.

However, rerankers still face significant challenges. As shown in Fig.~\ref{fig:motivation}, while they select the best passages based on relevance, this process often leads to redundancy and fails to capture complementary gains across passages. 
High-ranking passages may be redundant, wasting context budget and inference cost, whereas moderately relevant passages, when combined, can cover complementary aspects and add value. 
In short, a static top-k reranker has two issues: (i) it fails to fully explore the space of passage combinations, missing potential benefits; (ii) it often includes near-duplicate evidence, introducing unnecessary redundancy.

To address these challenges, we propose \textbf{OptiSet}, a set-centric framework that unifies set selection and set-level ranking in training. 
Specifically, we first propose an ``Expand-then-Refine'' paradigm, which first expands the query into multiple subqueries and then samples candidate passages to capture diverse perspectives. The resulting candidate pool is then further refined through re-selection, forming a complementary and compact set of evidence. 
Secondly, we use the perplexity variation of each set with respect to the generator as a utility signal, enabling the estimation of combinatorial gain. 
By combining the selection paradigm and utility estimation, we can construct high-quality training data without relying on strong LLM model supervision. 
Finally, we propose a set-list wise training strategy that jointly set selection and ranking, preserving the diversity of selection while avoiding the overfitting of surface patterns that is common in SFT training.

In summary, our contributions can be summarized as follows: 
\begin{itemize}
    \item We propose an ``Expand-then-Refine'' paradigm to construct compact and efficient evidence sets, together with a self-synthesis and labeling scheme for training data.
    \item We propose a set-list wise training strategy that jointly models set selection and set-level ranking in a unified framework.
    \item Extensive experiments demonstrate that our framework and training improve the performance of complex combinatorial problems and make the sets compact and efficient. 
\end{itemize}

\begin{figure*}[!ht]
    \centering
    \includegraphics[width=1\linewidth]{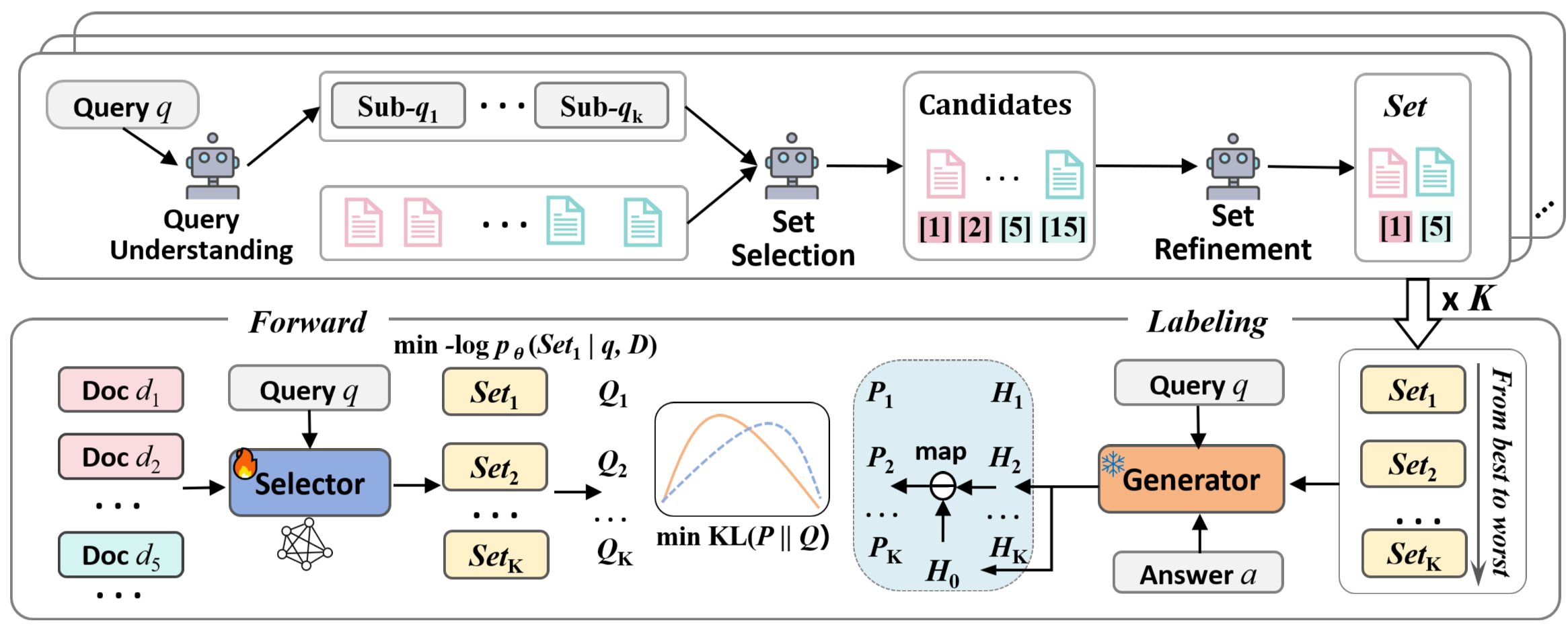}
    \caption{The framework diagram of OptiSet is shown. The upper part represents the ``Expand-then-Refine'' paradigm, which can serve as both a training-free framework and a training data synthesis pipeline. 
    The lower part represents set-list wise training. It utilizes multiple sets generated by the framework, performs partial ordering and signal labeling, and trains the selector.}
    \label{fig:framework}
\end{figure*}

\section{Related Works}
\subsection{Retrieval-Augmented Generation}



In recent years, RAG has emerged as a promising approach to mitigate limitations such as knowledge obsolescence and hallucinations~\citep{bechard2024reducing,gade2025s} in LLMs. It leverages externally retrieved knowledge to augment LLMs~\citep{lewis2020retrieval,zhao2024retrieval1,fan2024survey,li2025matching}.
Classical RAG systems typically follow a ``retrieve-then-read'' paradigm, first using a retriever and then employing a generator to produce responses~\citep{lewis2020retrieval}. However, this approach still faces numerous challenges that limit its effectiveness.

To address these challenges, recent research has focused on identifying retrieval intent to skip unnecessary searches~\citep{jeong2024adaptive,cheng2024unified,wang2023self} and optimizing queries to better align with retrievers~\citep{ma2023query,xu2024search,jiang2025qagent}. To mitigate noise and irrelevant passages, some methods introduce denoising~\citep{wei2025instructrag}, reranking~\citep{glass2022re2g}, selection~\citep{jiang2025gainrag,lee2025shifting,wang2025optimal}, and compression~\citep{xu2024recomp} between retrieval and generation.

\subsection{Reranker between Retrievers and LLMs.}



To bridge the gap between retrieval and generation, some studies introduce trainable middleware~\citep{jiang2025gainrag,dong2025understand}, particularly rerankers~\citep{glass2022re2g}. Recently, LLMs have been used as ranking models and have shown strong potential, becoming an active research direction~\citep{pradeep2023rankzephyr,weller2025rank1,zhuang2024setwise,liu2025reasonrank}. RankZephyr~\citep{pradeep2023rankzephyr} uses LLMs as a list-wise reranker to capture global passage relevance. RankR1~\citep{zhuang2025rank} and ReasonRank~\citep{liu2025reasonrank} leverage diverse reward designs and reinforcement learning to elicit LLMs' reasoning capabilities.




However, these methods often overlook the holistic nature of set-level evidence: passages can be complementary, and combining them can yield larger gains. SetR~\citep{lee2025shifting} takes an initial step via prompting and distillation fine-tuning. Nevertheless, it still relies on strong LLM supervision, and straightforward fine-tuning may overfit to surface patterns, limiting generalization. 

In contrast, we propose a self-synthesized paradigm that unifies set selection and ranking. Our approach achieves strong performance without added inference overhead, while substantially improving document selection efficiency.



\section{Method}
Our framework comprises two main components: the ``Expand-then-Refine'' paradigm and a set–list wise training framework. The paradigm generates diverse yet efficient candidate sets, while the training framework leverages these candidates to unify the set-level selection and ranking. 
This section details the core methods, as shown in Fig~\ref{fig:framework}.

\subsection{Expand-then-Refine for Set Selection}
The combinatorial selection space is extremely large, as the set size is not fixed. For an input of 20 documents, the space contains $2^{20}$ possible subsets. 
To find a globally better yet compact evidence set, we propose an ``Expand-then-Refine'' paradigm that leverages LLM reasoning: expand the query for broader exploration, then refine after an initial selection for targeted re-selection. This is training-free and also supports our data synthesis.

\subsubsection{Expansion: Query Understanding}
Complex problems often require multi-hop reasoning, and simply identifying the original query is insufficient to obtain comprehensive information. To alleviate this limitation, we introduce an expansion phase before set selection to achieve a thorough understanding of the original query. Specifically, in this phase, we prompt the LLM to reinterpret the original query $q$, generating multiple subqueries or variations, each capturing a different aspect of the problem. This process can be formalized as follows: 
\begin{equation}
\label{eq:Expand}
    Q_s = \text{Expand}(q) =  \{ q_1, q_2, \dots, q_k \}, 
\end{equation}
where \( q_1, q_2, \dots, q_k \) are generated by applying query decomposition (or rewriting) to the original query $q$. This provides multi-perspective views of the query, enabling subsequent stages to select candidate sets with more comprehensive coverage.

\subsubsection{Passage Set Selection}
\label{sec:selection}

After the expansion stage, inspired by CoT~\citep{wei2022chain}, we leverage step-by-step LLM reasoning for set selection, avoiding random sampling over the large combinatorial space. 
Concretely, following SetR~\citep{lee2025shifting}, we prompt the LLM to perform three reasoning steps: (1) list the key information required to answer the question and its sub-questions; (2) identify passages that contain each required item; (3) form a passage subset that provides the most comprehensive coverage. 

Putting the above steps together, the LLM selects multiple candidate passages for the original query \(q\) and each sub-query \(q_i \in Q_s\). For any query \(q'\in \{q\}\cup Q_s\), we denote its candidate passage set as
\(\mathcal{C}(q')=\{s_1,s_2,\dots,s_n\}\).
Our objective is to construct an evidence set \(\mathcal{S}\subseteq\mathcal{D}\) by aggregating candidates across all queries:
\(\mathcal{S}=\bigcup_{q' \in \{q\}\cup Q_s}\mathcal{C}(q')\).
We summarize this procedure as
\begin{equation}
\label{eq:Selection}
    \mathcal{S}_\text{raw} = \text{Select}(q, Q_s,\mathcal{D})
\end{equation}
This produces a diverse, high-coverage evidence pool by leveraging decomposition in the expansion stage and stepwise selection, and provides a strong starting point for refinement in the next stage.

\subsubsection{Refinement: Set Re-selection}
After generating a diverse set of candidate passages, \(\mathcal{S}_{\text{raw}}\) there may still contain redundant or irrelevant passages. To address this, we therefore apply a refinement step to filter such candidates, producing a smaller and more efficient set. 

Concretely, we repeat the selection procedure in \S\ref{sec:selection}, but replace the original candidate pool \(\mathcal{D}\) with \(\mathcal{S}_{\text{raw}}\) and drop the decomposed queries \(Q_s\), using only the original query \(q\) as a global constraint: 
\begin{equation}
\label{eq:Refinement}
    \mathcal{S}_\text{refined} = \text{Refine}(q, \mathcal{S}_\text{raw}).
\end{equation}
After refinement, we obtain \(\mathcal{S}_\text{refined}\), which contains fewer passages while preserving set-level relevance and complementary gains through LLM reasoning.

\subsection{Self-Synthesized Training Data}
We next describe how we self-synthesize training data for \textbf{OptiSet}. 
As summarized in Algorithm~\ref{alg:data_construction}, we run ``Expand-then-Refine'' multiple times to generate diverse evidence sets and label them with set-level utility. 

\subsubsection{Training Data Contruction}
By running the framework multiple times with high-temperature sampling, we can obtain multiple candidate sets for the same query. Formally,
\begin{equation}
    \mathcal{S} = [s_1, \dots,s_k] = \text{ESR}(q, \mathcal{D},[a]).
\end{equation} 
ESR($\cdot$) is a shorthand for the full ``Expand-then-Refine'' pipeline (i.e., Eq~\ref{eq:Expand},\ref{eq:Selection} and \ref{eq:Refinement}) and $[\cdot]$ represents optional input. 
Note that, unlike the training-free framework, training data construction uses the gold answer $a$ as an additional input, and its effect is detailed in \S~\ref{sec:ablation}.

\subsubsection{Signals Labeling}

\paragraph{Perplexity-based utility.}
To score each selected set, we use perplexity (PPL)~\citep{li2024quantity} as a dense utility signal; lower PPL indicates better evidence for generating the gold answer: 
\begin{equation}
\small{
    \text{PPL} = \exp\left(-\frac{1}{N} \sum_{j=1}^N \log p(a_{j} \mid q, S, a_{1}, \dots, a_{j-1})\right).
}
\end{equation}
To stabilize the scale of this signal, we convert PPL to entropy (i.e., log-perplexity), mathematically:
\begin{equation}
    H = \log(\text{PPL}).
\end{equation}
Thus, different selected sets can be compared via their utility under the generator. 


\paragraph{Adjusting Positive and Negative Samples}
As shown in prior work, a ``preference gap'' exists between retrievers and generators~\citep{ke2024bridging,jiang2025gainrag,dong2025understand}: low-entropy samples may still yield limited gains, since low entropy may reflect the generator’s internal knowledge and retrieved evidence does not necessarily help.
We therefore compute the baseline entropy (w/o passages) \(H_o\) and define
\begin{equation}
    \Delta H = H_p - H_o,
\end{equation}
where \(H_p\) is the entropy conditioned on the selected set. 
We treat sets with \(\Delta H \le 0\) as positive and those with \(\Delta H > 0\) as negative. 

To increase separability, we map \(\Delta H\) to a signed preference score:
\begin{equation}
\label{eq:map}
    P=
    \begin{cases}
    1-\operatorname{sigmoid}(\alpha \Delta H), & \Delta H \le 0,\\
    -\operatorname{sigmoid}(\beta \Delta H), & \Delta H > 0.
    \end{cases}
\end{equation}
Here, \(\alpha\) and \(\beta\) are scaling coefficients estimated from the data distribution (Appendix~\ref{sec:appendix-train}). 
After transformation, positive map to \(P \in (0.5, 1]\), while negative map to \(P \in [-1, -0.5)\).

\subsection{Set-List Wise Training}
To better leverage the sampled sets, we adopt a set-list wise training approach that jointly learns set selection and set-level ranking. The algorithm flow is summarized in Algorithm~\ref{alg:training}.

\paragraph{Supervised Learning}
To learn the optimal selection of the set, we take the highest gain score in the sampling group as an approximation of the optimal set selection and optimize it through loss on standard cross-entropy. 
\begin{equation}
\label{eq:argmax-s}
s^{*}=\arg\max_{s\in\mathcal{S}} \; P(a\mid q,s)
\end{equation}
\begin{equation}
\label{eq:ce}
\mathcal{L}_{\text{CE}}
= -\log p_\theta(s^{*}\mid q,\mathcal{D})
\end{equation}
Through optimal set fitting, the model will further enhance its set selection ability and make the output conform to the expected format. 

\paragraph{List-Wise Ranking}
In our training framework, we forward propagate multiple sets sampled from the same data, calculate their PPL values, and perform a softmax operation over the set scores. The softmax function normalizes the PPL values, assigning higher probabilities to better sets while retaining diversity by allowing for the possibility of selecting suboptimal sets. 
We first define the target distribution over sets based on their quality scores: 
\begin{equation}
    \label{eq:p_distribution}
    \mathcal{P} = \text{softmax}(\text{P}), \quad \mathcal{P}_i =  \frac{\exp(\text{P}_i)}{\sum_{j=1}^{m} \exp(\text{P}_j)}
\end{equation}
Next, we define the model-predicted distribution over the same set candidates. For each set $s_i$, the model assigns a sequence-level log-probability:
\begin{equation}
\ell_i = \log p_\theta(s_i \mid q, \mathcal{D}),
\end{equation}
and the predicted distribution is obtained via softmax normalization:
\begin{equation}
\label{eq:q_distribution}
\mathcal{Q} = \operatorname{softmax}(\boldsymbol{\ell}),
\quad
\mathcal{Q}i = \frac{\exp(\ell_i)}{\sum_{j=1}^{m}\exp(\ell_j)}
\end{equation}
This step is followed by the calculation of the KL-divergence between the predicted distribution and the true distribution of the sets:
\begin{equation}
\label{eq:kl}
\mathcal{L}_\text{KL} = KL(\mathcal{P}||\mathcal{Q}) = \sum_{i=1}^{m} \mathcal{P}_i \log \left( \frac{\mathcal{P}_i}{\mathcal{Q}_i} \right)
\end{equation}
This loss function allows the model to learn to rank sets by their quality while preserving the diversity within the sampled sets.

\paragraph{Objective}
Our final training involves combining the two losses together, formally, 
\begin{equation}
\label{eq:loss}
    \mathcal{L}_\text{all} = \mathcal{L}_\text{CE} + \lambda\mathcal{L}_\text{KL},
\end{equation}
where $\lambda$ is the balance coefficient.

\begin{algorithm}[!h]
\caption{Training Data Construction}
\label{alg:data_construction}
\textbf{Input:} Query $q$, gold answer $a$, corpus $\mathcal{D}$, sampling times $K$ \\
\textbf{Output:} Candidate set list $\mathcal{S} = \{s_1,\dots,s_K\}$ with quality signals $\mathbf{P}$. 
\begin{algorithmic}[1]
    \State Compute baseline $\mathrm{PPL}_0$ with empty set
    \State $H_0 \gets \log(\mathrm{PPL}_0)$ \Comment{Baseline entropy}

    \For{$k = 1$ to $K$}
        \State $Q_s^{(k)} \gets \text{Expand}(q,a)$ 
        \State $s_{raw}^{(k)} \gets \text{Select}(q, Q_s^{(k)},\mathcal{D}, a)$ 
        \State $s^{(k)} \gets \text{Refine}(q, s_{raw}^{(k)},a)$ \
        \State Compute $\mathrm{PPL}_k$ conditioned on $(q, s_k,a)$
        \State $H_k \gets \log(\mathrm{PPL}_k)$
        \State $\Delta H_k \gets H_k - H_0$ \Comment{Entropy difference}
        \State $P_k \gets \text{Transform}(\Delta H_k)$ \Comment{Eq.~\eqref{eq:map}}
    \EndFor
    \State \Return $\mathcal{S}=\{s_1,\dots,s_K\}$, $\mathbf{P}=\{P_1,\dots,P_K\}$
\end{algorithmic}
\end{algorithm}

\begin{algorithm}[!h]
\caption{Set-List Wise Training}
\label{alg:training}
\textbf{Input:} Training data $(q, \mathcal{D}, \mathcal{S}, \mathbf{P})$, model $p_\theta$ \\
\textbf{Output:} Updated model parameters $\theta$
\begin{algorithmic}[1]
    \For{each training instance $(q, \mathcal{D}, \mathcal{S})$}

        \State $s^* \gets \arg\max_{s_i \in \mathcal{S}} P_i$ \Comment{Eq.~\eqref{eq:argmax-s}}
        \State $\mathcal{L}_{\mathrm{CE}} \gets -\log p_\theta(s^* \mid q, \mathcal{D})$ \Comment{Eq.~\eqref{eq:p_distribution}}

        \For{each set $s_i \in \mathcal{S}$}
            \State $\ell_i \gets \log p_\theta(s_i \mid q, \mathcal{D})$
        \EndFor
        \State $\mathcal{Q} \gets \operatorname{softmax}(\boldsymbol{\ell})$ \Comment{Model distribution}

        \State $\mathcal{P} \gets \operatorname{softmax}(\mathbf{P})$ \Comment{Target distribution}
        \State $\mathcal{L}_{\mathrm{KL}} \gets \mathrm{KL}(\mathcal{P}\|\mathcal{Q})$ \Comment{Eq.~\eqref{eq:kl}}

        \State $\mathcal{L} \gets \mathcal{L}_{\mathrm{CE}} + \lambda \mathcal{L}_{\mathrm{KL}}$ \Comment{Eq.~\eqref{eq:loss}}
        \State Update $\theta$ using $\nabla_\theta \mathcal{L}$
    \EndFor
\end{algorithmic}
\end{algorithm}

\begin{table*}[!ht]
\centering
\small 
\renewcommand{\arraystretch}{1.1}
\setlength{\tabcolsep}{4.2pt} 

\resizebox{0.935\textwidth}{!}{%
\begin{tabular}{l ccc ccc ccc ccc}
\toprule
\multirow{2}{*}{\textbf{Method}} & \multicolumn{3}{c}{\textbf{HotpotQA}} & \multicolumn{3}{c}{\textbf{Bamboogle}} & \multicolumn{3}{c}{\textbf{MuSiQue}} & \multicolumn{3}{c}{\textbf{TriviaQA}} \\
\cmidrule(lr){2-4} \cmidrule(lr){5-7} \cmidrule(lr){8-10} \cmidrule(lr){11-13}
 & EM & F1 & doc & EM & F1 & doc & EM & F1 & doc & EM & F1 & doc \\
\midrule

\rowcolor{gray!8} \multicolumn{13}{c}{\textit{Zero-shot Baselines \& Ours Training-Free}} \\ \addlinespace[0.5ex]

Vanilla   & 25.20 & 26.88 & 0.00 & 13.60 & 18.18 & 0.00 & 6.54  & 10.14 & 0.00 & 63.60 & 64.40 & 0.00 \\
NaiveRAG        & 33.20 & 35.13 & 3.00 & 17.60 & 23.83 & 3.00 & 8.44  & 13.55 & 3.00 & 69.00 & 71.14 & 3.00 \\

Ours$^\dagger$         & \underline{38.20} & 41.72 & 2.20 & 22.40 & 29.08 & 2.30 & \underline{10.84} & {15.45} & 2.13 & 72.40 & 72.58 & 2.93 \\
{\color[HTML]{999999} SetR$_\text{O}$} & {\color[HTML]{999999} \textit{37.00}} & {\color[HTML]{999999} \textit{41.11}} & {\color[HTML]{999999} \textit{3.45}} & {\color[HTML]{999999} \textit{20.00}} & {\color[HTML]{999999} \textit{27.58}} & {\color[HTML]{999999} \textit{4.05}} & {\color[HTML]{999999} \textit{10.76}} & {\color[HTML]{999999} \textit{16.16}} & {\color[HTML]{999999} \textit{3.91}} & {\color[HTML]{999999} \textit{74.20}} & {\color[HTML]{999999} \textit{73.93}} & {\color[HTML]{999999} \textit{4.44}} \\

\addlinespace[0.8ex]
\rowcolor{gray!8} \multicolumn{13}{c}{\textit{Fine-tuned Baselines \& Ours}} \\ \addlinespace[0.5ex]

ReasonIR        & 37.40 & 40.22 & 3.00 & \underline{24.80} & 29.85 & 3.00 & 10.47 & 14.94 & 3.00 & 72.60 & 73.43 & 3.00 \\
RankZephyr      & 33.60 & 39.19 & 3.00 & 20.80 & 28.69 & 3.00 & 10.76 & 15.36 & 3.00 & 72.80 & 73.05 & 3.00 \\
SetWise         & 35.40 & 40.41 & 3.00 & 21.60 & 28.01 & 3.00 & 9.64  & 14.86 & 3.00 & \underline{74.20} & 74.41 & 3.00 \\
RankR1          & 38.00 & \underline{42.23} & 3.00 & 23.20 & 29.72 & 3.00 & 10.30 & 15.33 & 3.00 & 73.00 & 73.40 & 3.00 \\
Rank1           & 35.20 & 40.09 & 3.00 & 22.40 & \underline{30.60} & 3.00 & 10.51 & \textbf{15.84} & 3.00 & \underline{74.20} & \underline{74.61} & 3.00 \\
Ours & \textbf{38.80} & \textbf{43.30} & 2.34 & \textbf{25.60} & \textbf{32.67} & 2.47 & \textbf{10.84} & \underline{15.49} & 2.47 & \textbf{75.40} & \textbf{75.83} & 2.70 \\ 
\bottomrule
\end{tabular}
}
\caption{EM/F1(\%) of different methods experimented on four datasets. The best and second best scores are highlighted in \textbf{bold} and \underline{underlined}, respectively. {\color[HTML]{999999}\textit{Gray}} is not used for comparison. Ours$^\dagger$ donate Training-free.}
\label{tab:main}
\end{table*}

\section{Experiments}

\subsection{Implementation Details}
\paragraph{Training Data.} 
We sample a subset of the training split from HotpotQA datasets, running our framework 10 times per question. After collecting 20K samples, we apply an initial filter to remove examples that are entirely incorrect or lack discriminative signal. To reduce \textbf{bias from passage IDs} in the training distribution, we shuffle the the order of the input passage. \textbf{three} times per query. Finally, we retain 5 sets per sample and randomly subsample 8K samples for training. 

\paragraph{Training Details.} Training Details We fine-tune LLaMA3.1-8B with LoRA (r=128, $\alpha$=32, dropout=0.05). During SFT, we train for 1 epoch with a learning rate of 3e-5. The loss balancing factor $\lambda$ is set to 0.1. All experiments are
conducted on 2$\times$ A100 GPUs. 

\paragraph{Inference Details.}  During inference, we use Contriever~\citep{izacard2021unsupervised} as the retriever and set k to 20. For all datasets, we use 21M English Wikipedia dump as the source passages for the retrieval~\citep{karpukhin2020dense}. Prompts for the experiments can be found in Appendix~\ref{sec:appendix-prompts}.

\subsection{Datasets and Evaluation}
\paragraph{Datasets. }
We evaluate our OptiSet on four open-domain QA datasets covering both multi-hop and single-hop. The multi-hop QA benchmarks include HotpotQA~\citep{yang2018hotpotqa}, Bamboogle~\citep{press2023measuring} and MuSiQue~\citep{trivedi2022musique}. For general QA, we use TriviaQA~\citep{joshi2017triviaqa}. We adopt 500-sample subsets of 2WikiMultiHopQA, HotpotQA and TriviaQA for efficiency, following ~\citep{trivedi2022interleaving,kim2024sure}. 

\paragraph{Evaluation.} 
We report two metrics: EM and F1. Following prior work~\citet{asai2023self,mallen2022not,jiang2025cocoa}, we use a non-strict EM where a prediction is correct if it contains the gold answer. F1 measures token-level overlap between the predicted and gold answers. Since longer response may improve EM via coverage but introduce noise that lowers F1, evaluating both metrics allows for a more balanced evaluation.

\subsection{Baselines}
We select several of the most representative methods for comparison. 
(1) Vanilla: direct answering without retrieval.
(2) Naive RAG, which is the classic “retrieve-then-read” paradigm.
(3) SetR$_\text{O}$~\citep{lee2025shifting}: a zero-shot version of set selection based on LLM reasoning.
(4) ReasonIR~\citep{shao2025reasonir}: a powerful 7B retriever with strong inferential retrieval capabilities. 
(5) RankZephyr~\citep{pradeep2023rankzephyr}: The classic approach is to use LLM for list-wise ranking.
(6) SetWise~\citep{zhuang2024setwise}: a setwise algorithm that uses heap sort to produce a full ranking.
(7) RankR1~\citep{zhuang2025rank}: a reasoning reranker trained with RL.
(8) Rank1~\citep{weller2025rank1}: reranking with increased test-time computation.
All retrieval-based methods use top-20 passages. 
All reranker only selected the top-3 evidence.
Details experimental settings are shown in the Appendix~\ref{sec:appendix-dataset} and \ref{sec:appendix-baselines}. 

\subsection{Main Results}
In our experiments as shown in Table~\ref{tab:main},  we summarize several key findings:

\textbf{(1) The role of re-ranking or selection.}
Adding retrieval with re-ranking consistently improves over direct answering and naive retrieval.
Compared with \textit{NaiveRAG}, our method improves EM/F1 by +5.00/+6.59 on HotpotQA and by +4.80/+5.25 on Bamboogle in the zero-shot setting, demonstrating that the improvement comes from effectively selecting high-quality evidence.

\textbf{(2) Reasoning vs.\ Non-reasoning.}
We observe that rerankers with explicit reasoning capabilities generally outperform non-reasoning approaches. 
Methods such as \textit{RankR1} achieve higher EM/F1 than non-reasoning ranking baselines (e.g., \textit{RankZephyr}) on most datasets. 
We attribute this advantage to cognitive chaining, enabling the model to identify the intermediate steps required for multi-hop reasoning and better assess passage relevance.
Notably, our method remains competitive without explicit reasoning, indicating the effectiveness of our selection mechanism.

\textbf{(3) Expand-then-Refine vs.\ SetR.}
SetR$_\text{O}$ is competitive, suggesting reasoning-based selection is effective, but it tends to over-select passages, increasing cost.
Our expand-then-refine paradigm achieves higher accuracy with fewer documents (e.g., Bamboogle: +2.40 EM with 2.30 vs.\ 4.05 \texttt{doc}).
Overall, refinement is crucial for mitigating over-selection and improving evidence efficiency.

\textbf{(4) Efficiency.}
Our approach improves both generation and training efficiency.
It selects fewer passages than prior rerankers, reducing context length and generation inference cost, and it avoids reinforcement learning, simplifying training and improving stability without sacrificing performance.

\begin{table*}[t]
\centering
\small
\setlength{\tabcolsep}{1.0pt} 
\renewcommand{\arraystretch}{1.25}
\begin{tabularx}{\textwidth}{l @{\extracolsep{\fill}} cccc cccc cccc c}
\toprule
\multirow{2}{*}{\textbf{Method}} & \multicolumn{4}{c}{\textbf{HotpotQA}} & \multicolumn{4}{c}{\textbf{Bamboogle}} & \multicolumn{4}{c}{\textbf{MuSiQue}} & \multirow{2}{*}{\textbf{Avg.}} \\
\cmidrule(lr){2-5} \cmidrule(lr){6-9} \cmidrule(lr){10-13}
 & EM & F1 & Avg & doc & EM & F1 & Avg & doc & EM & F1 & Avg & doc & \\
\midrule

\rowcolor{gray!10} \multicolumn{14}{c}{\textit{Pipeline Ablation: Effectiveness of Different Modules}} \\ \addlinespace[0.5ex]
Selection-only  & 37.00 & 41.11 & 39.06 & 3.45 & 20.00 & 27.58 & 23.79 & 4.05 & 10.76 & \textbf{16.16} & \textbf{13.46} & 3.91 & 25.44 \\
$w/o$ Expand     & 36.60 & 40.84 & 38.72 & \textbf{1.99} & \underline{20.80} & \underline{28.64} & \underline{24.72} & \textbf{2.07} & 10.63 & 15.89 & 13.26 & \textbf{2.02} & 25.57 \\
$w/o$ Refine     & \textbf{39.00} & \textbf{42.13} & \textbf{40.57} & 4.45 & 19.20 & 26.56 & 22.88 & 5.22 & 10.47 & 15.76 & 13.12 & 4.51 & 25.52 \\
$w/$  All  & \underline{38.20} & \underline{41.72} & \underline{39.96} & \underline{2.20} & \textbf{22.40} & \textbf{29.08} & \textbf{25.74} & \underline{2.30} & \textbf{10.84} & \underline{15.45} & \underline{13.15} & \underline{2.13} & \textbf{26.28} \\
\textit{\color{gray}Ours+Answer} & \textit{\color{gray}39.00} & \textit{\color{gray}43.64} & \textit{\color{gray}41.32} & \textit{\color{gray}2.09} & \textit{\color{gray}23.20} & \textit{\color{gray}30.67} & \textit{\color{gray}26.94} & \textit{\color{gray}1.78} & \textit{\color{gray}13.20} & \textit{\color{gray}17.91} & \textit{\color{gray}15.55} & \textit{\color{gray}1.95} & \textit{\color{gray}27.94} \\

\midrule \addlinespace[0.5ex]

\rowcolor{gray!10} \multicolumn{14}{c}{\textit{Training Ablation: Comparison of Training Strategies}} \\ \addlinespace[0.5ex]
SFT            & 32.80 & 36.76 & 34.78 & \textbf{1.95} & 20.80 & 26.52 & 23.66 & \textbf{1.94} & 7.57 & 12.18 & 9.88 & \textbf{1.94} & 22.77 \\
DPO            & \underline{37.40} & \underline{41.76} & \underline{39.58} & 2.87 & \underline{20.80} & \underline{28.73} & \underline{24.77} & 3.19 & \underline{10.67} & \underline{15.18} & \underline{12.93} & 3.13 & \underline{25.76} \\
Ours  & \textbf{38.80} & \textbf{43.30} & \textbf{41.05} & \underline{2.34} & \textbf{25.60} & \textbf{32.67} & \textbf{29.13} & \underline{2.47} & \textbf{10.84} & \textbf{15.49} & \textbf{13.16} & \underline{2.47} & \textbf{27.78} \\
\bottomrule
\end{tabularx}
\caption{Ablation studies on both pipeline modules and training strategies. \textbf{Bold} and \underline{underline} denote the best and second-best results in each group respectively. {\color[HTML]{999999}\textit{Gray}} is not used for comparison.}
\label{tab:combined-ablation}
\end{table*}

\subsection{Ablation Study I: Paradigm}
\label{sec:ablation}
We ablate the expand-then-refine paradigm by removing each module while keeping all other settings fixed (top part of Table~\ref{tab:combined-ablation}). 


\textbf{Effect of \textit{Refinement}.}
\textit{Refine} is the main driver of efficiency.
Removing \textit{Expand} (\emph{w/o Expand}) yields the most compact evidence sets (about 2\texttt{doc}) with comparable performance to other variants, indicating that refinement alone can perform strong selection under a tight budget.

\textbf{Effect of \textit{Expansion}.}
Using \textit{Expand} alone is not consistently beneficial: \emph{w/o Refine} improves HotpotQA  but degrades other datasets while selecting substantially more passages.
In contrast, \textit{Expand} complements \textit{Refine}: compared with \emph{w/o Expand}, the full paradigm (\emph{All}) yields higher performance (e.g., 25.57$\rightarrow$26.28) with only a marginal increase in evidence size.
These results suggest that expansion broadens candidate coverage, allowing \textit{Refine} to select better evidence than refinement alone.

\textbf{Effect of \textit{Answer}.}
During data synthesis, we additionally condition on the gold answer as guidance. This consistently resulted in higher performance, further improving the quality of the training data.


\subsection{Ablation Study II: Training}
We further compare training strategies for the selector under the same evaluation protocol (Table~\ref{tab:combined-ablation}).

\textbf{SFT vs.\ DPO.}
SFT produces the smallest evidence sets but substantially hurts accuracy, suggesting it may learn overly simple selection patterns. 
DPO improves accuracy over SFT but selects more passages and remains suboptimal. 

\textbf{Set-List wise training.}
Our training achieves the best overall performance with a compact evidence size.
It consistently outperforms DPO across datasets, with particularly large gains on Bamboogle (Avg 29.13 vs.\ 24.77) while using fewer passages (2.47 vs.\ 3.19 \texttt{doc}).
These results suggest that set--list-wise supervision improves both accuracy and evidence efficiency compared to point-wise SFT and pair-wise DPO.

\begin{figure}[!ht]
    \centering
    \includegraphics[width=0.995\linewidth]{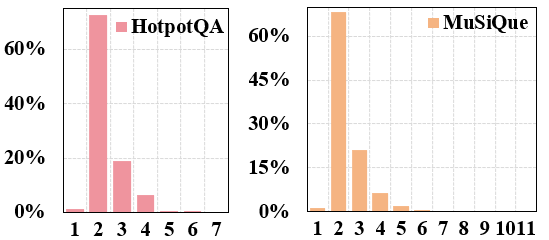}
    \caption{Illustration of the distribution of the number of documents dynamically selected from a set.}
    \label{fig:distribution}
\end{figure}



\subsection{Dynamic Set Distribution and Efficiency}
We analyze the distribution of selected set sizes to characterize the behavior of dynamic evidence selection.
As shown in Fig.~\ref{fig:distribution}, most questions are answered with compact evidence sets of 2-3 passages, indicating strong efficiency.
Meanwhile, the method occasionally selects larger sets, suggesting it can adaptively increase coverage when additional information is required.
This automatic trade-off, however, reduces user control over the evidence budget, which is a limitation.




\subsection{Redundancy Analysis}
To quantify redundancy reduction, we follow \citet{clarke2008novelty,tsai2010evaluation} and report $Novel$ (see Appendix~\ref{sec:appendix-redundancy} for details), a novelty-aware metric that penalizes repetition after accounting for correlation.

Table~\ref{tab:novel} shows that our method achieves high overall novelty when averaged across all examples, partly because it selects fewer passages on average; importantly, selecting fewer passages is itself a desirable form of redundancy reduction.
To control for set size, we further compare the most frequent cases with 2- and 3-passage sets against \textit{RankZephyr}.
Under matched sizes, our novelty remains higher than the classic list-wise ranker, indicating room for improving redundancy removal beyond compact selection.

\begin{table}[t]
\centering
\small
\setlength{\tabcolsep}{3pt}
\renewcommand{\arraystretch}{1.15}

\setlength{\aboverulesep}{0.75pt} 
\setlength{\belowrulesep}{0.75pt} 

\begin{tabular}{@{} l l S[table-format=2.2] S[table-format=2.2] S[table-format=2.2] @{}}
\toprule
\textbf{Metric} & \textbf{Model} & \textbf{HotpotQA} & \textbf{Bamboogle} & \textbf{MuSiQue} \\
\midrule

\rowcolor{gray!12}
\multicolumn{5}{@{}l@{}}{\textbf{Jaccard}} \\
\multirow{2}{*}{Novel@all} & RankZep & 79.94 & 84.49 & 84.43 \\
                           & Ours    & \textbf{84.28} & \textbf{87.21} & \textbf{87.48} \\
\cmidrule{2-5}
\multirow{2}{*}{Novel@2}   & RankZep & 81.80 & 87.90 & 86.55 \\
                           & Ours    & \textbf{85.56} & \textbf{88.92} & \textbf{88.56} \\
\cmidrule{2-5}
\multirow{2}{*}{Novel@3}   & RankZep & 79.17 & \textbf{85.28} & 84.51 \\
                           & Ours    & \textbf{80.58} & 85.03 & \textbf{84.97} \\
\midrule

\rowcolor{gray!12}
\multicolumn{5}{@{}l@{}}{\textbf{BERTScore F1}} \\
\multirow{2}{*}{Novel@all} & RankZep & 56.34 & 58.96 & 59.08 \\
                           & Ours    & \textbf{64.76} & \textbf{65.60} & \textbf{66.41} \\
\cmidrule{2-5}
\multirow{2}{*}{Novel@2}   & RankZep & 65.57 & 68.90 & 68.40 \\
                           & Ours    & \textbf{67.68} & \textbf{69.71} & \textbf{69.72} \\
\cmidrule{2-5}
\multirow{2}{*}{Novel@3}   & RankZep & 55.94 & 59.40 & 59.27 \\
                           & Ours    & \textbf{56.74} & \textbf{59.53} & \textbf{59.62} \\
\bottomrule
\end{tabular}

\caption{Passage novelty comparison between RankZephyr and our method.}
\label{tab:novel}
\end{table}

\begin{figure}[!b]
    \centering
    \includegraphics[width=0.995\linewidth]{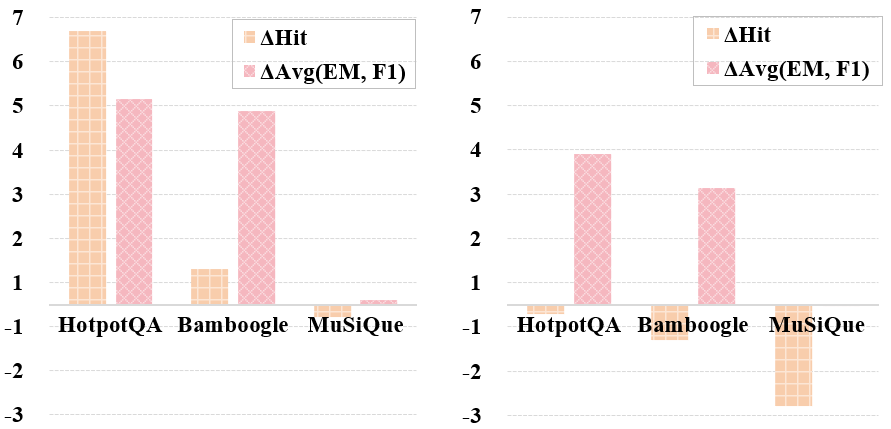}
    \caption{Illustration of gain. Changes in hit of the gold answer and QA performance using OptiSet. The left side compares to RankZephyr, and the right side compares to Rank1. Avg(EM, F1) is the metric.}
    \label{fig:gain}
\end{figure}

\subsection{What Gains does Self-Labeling bring?}
To verify whether our gains mainly come from higher passage recall, we compare the change in hit rate between OptiSet and other reranking baselines.

As shown in Figure~\ref{fig:gain}, compared to \textit{RankZephyr} and \textit{Rank1}, we rarely observed significant increases in OptiSet's hit rate, except for Hotpot; most showed only minor changes, or even a decrease in hit rate.
Nevertheless, answer quality still improved in these cases.
Since the generator was not trained, these improvements cannot be explained by generator adaptation, suggesting that the improvements stem from selecting higher-quality (but not necessarily higher recall) evidence sets. We hypothesize that the performance improvement is due to the combined gain from different passages selected by OptiSet, rather than from individual gold passages. 
However, different generators may have different document combination preferences, and improving generalization is a future direction.




\begin{figure}[!h]
    \centering
    \includegraphics[width=1.0\linewidth]{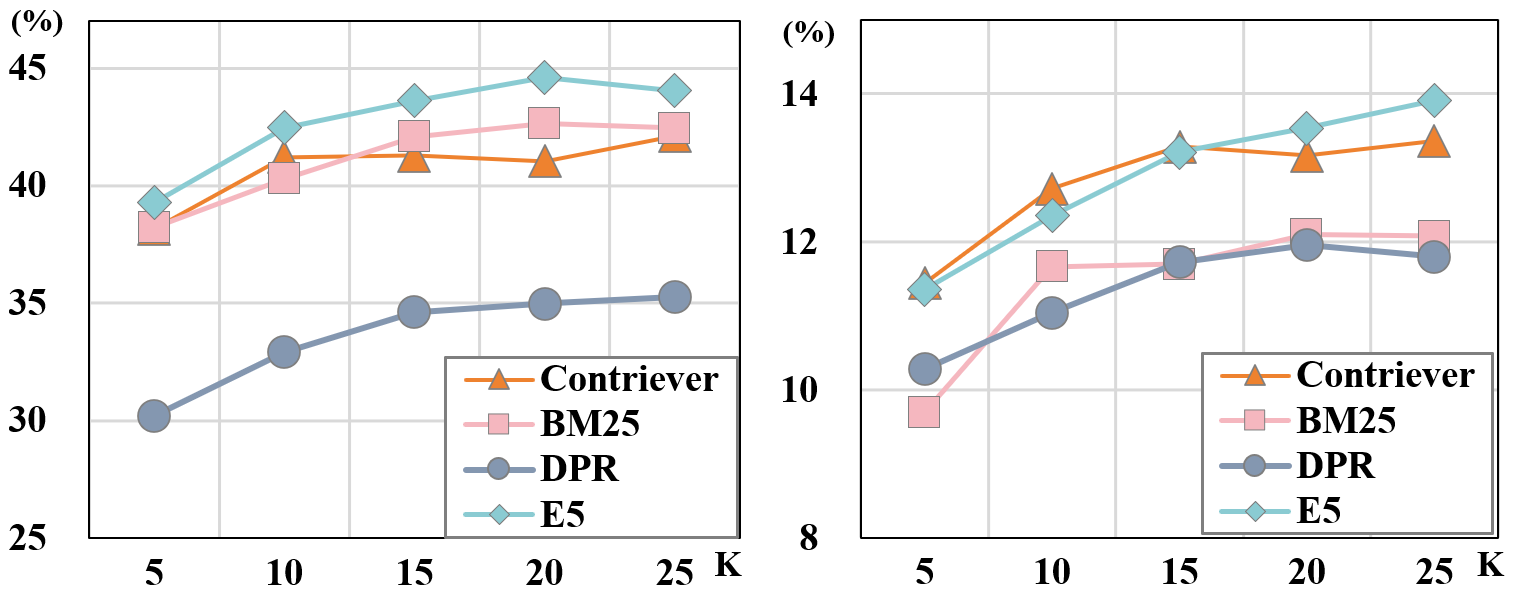}
    \caption{Illustration showing how generation performance changes as the candidate sizes changes or different retrievers are replaced. The left part is HotpotQA, and the right part is MuSiQue. For fairness, Avg(EM, F1) is uesd as the evaluation metric.}
    \label{fig:K}
\end{figure}

\subsection{Robustness to Different Retrievers and K}
To evaluate the generalization ability of our \textbf{optiSet} to different candidate sizes and its robustness to different retrieval systems, we observed the following:

\textbf{Retriever robustness.}
OptiSet performs consistently across retrieval systems, achieving strong results with both dense (e.g., E5) and sparse (BM25) retrievers.
This suggests that our selector is largely retriever-agnostic and does not overfit to a specific candidate distribution induced by the training setup.

\textbf{Sensitivity to K.}
Varying K follows the expected information-coverage trend: larger candidate pools generally improve accuracy by providing more relevant evidence to choose from.
Importantly, increasing K beyond the training setting (e.g., K{=}25) still yields gains, indicating that \textbf{optiSet} generalizes beyond the training-time candidate-size bias.




\section{Conclusion}
In this work, we investigate the combinatorial selection problem and propose OptiSet, 
a unified framework for set selection and set-level ranking in retrieval-augmented generation. OptiSet uses an \emph{expand-then-refine} paradigm to construct compact evidence sets and a self-synthesis scheme to label training data. We further introduce set-listwise training that jointly optimizes set selection and set-level ranking. Experiments show that OptiSet improves both efficiency and performance.

\section*{Limitations}
Despite achieving impressive performance, our approach still has some limitations in certain aspects: 
\begin{itemize}
    \item LLM-based methods are limited by context length; how to accept more candidates is a key research direction for us in the future. 
    \item Many RL baselines have achieved better performance. Although we achieved excellent performance without RL, how to stimulate inference ability through RL is still worth exploring.
    \item While the selection is dynamic top-k, which increases intelligence but sacrifices flexibility, incorporating K-budget into the instructions is a very worthwhile area of research.
\end{itemize}

\bibliography{custom}


\appendix
\section{Dataset}
\label{sec:appendix-dataset}
Here, we introduce in detail the datasets we used,
which are four datasets on two tasks

\textbf{HotpotQA}~\citep{yang2018hotpotqa}: It is a multi-hop question answering dataset based on Wikipedia. Given the cost constraints of the experiment, we used a subsample of the \citet{trivedi2022interleaving} dataset, which was obtained by extracting 500 questions from the validation set of each dataset.

\textbf{Bamboogle}~\citep{press2023measuring}: Bamboogle is a small multi-hop question-answering benchmark suite developed to study the ``combinatorial reasoning gap'' in language models. It contains 125 questions. Each article is manually written out, requiring two hops of information to answer. It deliberately retains questions where "evidence can be found on Wikipedia, but searching for the question directly on mainstream search engines often fails to yield the correct answer/the summary will be incorrect."

\textbf{MuSiQue}~\citep{trivedi2022musique}: By iteratively selecting composable single-hop question pairs, we created questions spanning 2-4 hops, which are generally more difficult.

\textbf{TriviaQA}~\citep{joshi2017triviaqa}: A compilation
of trivia questions paired with answers, both originally pulled from online sources. For the sake of inference efficiency, we randomly sample 500 entries here.

\begin{table}[!ht]
\centering
\setlength{\tabcolsep}{6pt}        
\renewcommand{\arraystretch}{1.15} 

\begin{tabularx}{0.875\linewidth}{@{} l l >{\raggedleft\arraybackslash}X @{}}

\toprule
\textbf{Task Type}    & \textbf{Dataset}     & \textbf{\# Samples} \\ 
\midrule
\multirow{3}{*}{Multi-HopQA} & HotpotQA & 500    \\
                            &  Bamboogle  & 125    \\
                            &  MuSiQue  & 2,417    \\
\midrule
One-HopQA      & TriviaQA       & 500   \\
\bottomrule
\end{tabularx}
\caption{Description of tasks and evaluation datasets.}
\label{tab:datasets}
\end{table}

\section{Baselines}
\label{sec:appendix-baselines}
We selected several of the most representative methods for comparison. 
\begin{itemize}[label={}]
    \item (1) \textbf{Vanilla}: Direct answer generation without any external retrieval. The model relies solely on its parametric knowledge (and the given prompt) to produce an answer, serving as a closed-book baseline that isolates the effect of adding retrieval.
    \item (2) \textbf{Naive RAG}: The standard ``retrieve-then-read'' pipeline. A retriever first fetches top-k passages, and the generator then conditions on the concatenated retrieved evidence to produce the final answer.

    \item (3) \textbf{SetR}: Leveraging the model's powerful reasoning capabilities, it performs set selection through long-chain cognition. Its fine-tuned version requires strong model supervision; since weights are not open, we compare it with a zero-shot version.
    
    \item (4)\textbf{ ReasonIR}~\citep{shao2025reasonir}: ReasonIR trains a strong retriever with a synthetic data pipeline that generates reasoning-required queries and “plausible-but-useless” hard negatives for each document. Due to its large capacity and strong performance, we use it as a representative strong retriever and employ it as the reranker used to supply candidates for reranking.
    
    \item (5) \textbf{RankZephyr}~\citep{pradeep2023rankzephyr}: A representative listwise LLM reranker built on Zephyr-7B. It is instruction-distilled from GPT-4/RankGPT-style teacher signals to directly output an ordered list of document indices for a candidate set.
    
    \item (6) \textbf{SetWise}~\citep{zhuang2024setwise}: A cost-efficient zero-shot reranking strategy combining setwise prompting with a sorting procedure. Instead of pairwise comparisons, the LLM selects the most relevant item from a small set of candidates each time, and a heap-sort-style process aggregates these selections into a full ranking with fewer LLM calls.
    
    \item (7) \textbf{Rank-R1}~\citep{zhuang2025rank}: A reasoning-enhanced reranker that uses setwise prompting plus reinforcement learning (e.g., GRPO). It improves “think-then-rank” behavior using only relevance supervision, rather than relying solely on SFT or prompting.
    
    \item (8) \textbf{Rank1}~\citep{weller2025rank1}: A reranking approach that brings test-time compute to ranking via reasoning-trace distillation. It curates and releases 600K+ reasoning traces from strong reasoning models (e.g., DeepSeek-R1) on MS MARCO and distills them into a smaller reranker that generates a reasoning trace before producing relevance judgments/rankings. 
\end{itemize}
All retrieval-based methods use top-20 passages. 
And all reranker only selected the top-3 evidence. 

\section{Training \& Inference Details}
\label{sec:appendix-train}
\paragraph{Training Data.} We primarily sampled 20k data poinsts from training dataset of HotpotQA~\citep{yang2018hotpotqa}, running the framework 10 times per question. 
After obtaining 20k data points, we performed initial filtering, removing those that were completely erroneous or lacked significant discriminative power. During training, each data point was kept with 5 ranking candidates. Additionally, to reduce the bias in paragraph IDs within the training data distribution, we randomly shuffled the input data three times (equivalent to 3 epochs). Finally, our preliminary experiments showed that convergence could be achieved with only partial training, so we randomly sampled 8k data points for analysis.

\paragraph{Training Details.} Training Details We fine-tune LLaMA3.1-8B with LoRA (r=128, $\alpha$=32, dropout=0.05). During SFT, we train for 1 epoch (he data has been shuffled 3 times during construction) with a learning rate of 3e-5. All experiments are
conducted on 2$\times$ A100 GPU. 

\paragraph{Inference Details.}  During inference, we use Contriever~\citep{izacard2021unsupervised} as the retriever and set k to 20. For all datasets, we use 21M English Wikipedia dump as the source passages for the retrieval~\citep{karpukhin2020dense}. 
In addition, we also used a series of other retrievers in the analysis section, including BM25~\citep{robertson1994some}, DPR~\citep{karpukhin2020dense}, and E5~\citep{wang2022text}. 
Furthermore, we use the vllm\footnote{https://docs.vllm.ai/en/latest/} library to speed up inference.

\paragraph{Evaluation} 
We report two metrics: EM and F1. Following prior work~\citet{asai2023self,mallen2022not}, we use a non-strict EM where a prediction is correct if it contains the gold answer. F1 measures token-level overlap between the predicted and gold answers. Since longer response may improve EM via coverage but introduce noise that lowers F1, evaluating both metrics allows for a more balanced evaluation. 
In addition, to measure efficiency, we reported the documents used after the reranking.

\paragraph{Fitting $\alpha$ and $\beta$.}
We estimate $(\alpha,\beta)$ by minimizing a goodness-of-fit loss on the resulting score distributions, using the sum of Kolmogorov--Smirnov distances (negative scores vs.\ $\mathrm{Unif}[-1,-0.5]$ and positive scores vs.\ $\mathrm{Unif}[0.5,1]$), with $\alpha,\beta\in[0.01,10]$ optimized via \texttt{scipy.optimize.minimize}.

\section{Novelty Metrics}
\label{sec:appendix-redundancy}

We evaluate redundancy removal using a novelty indicator adapted from novelty-based evaluation~\citep{clarke2008novelty,tsai2010evaluation}. 
Given the selected passages for a query in ranked order $\mathcal{P}=\langle p_1,\ldots,p_n\rangle$, we compute the marginal novelty of each passage as
\begin{equation}
g_1=1,\qquad 
g_i = 1-\max_{j<i}\mathrm{Sim}(p_i,p_j)\ (i\!>\!1),
\end{equation}
where $\mathrm{Sim}(\cdot,\cdot)\in[0,1]$ measures passage similarity (we use Jaccard or BERTScore-F1).
The novelty of the set is the average marginal gain:
\begin{equation}
\mathrm{Novel}(\mathcal{P})=\frac{1}{n}\sum_{i=1}^{n} g_i .
\end{equation}
This matches our implementation: the first passage is treated as fully new, and each subsequent passage contributes $1$ minus its maximum similarity to earlier selections.

\paragraph{Novel@all / Novel@2 / Novel@3.}
We report dataset-level novelty by averaging $\mathrm{Novel}(\mathcal{P})$ over queries. 
\textbf{Novel@all} averages over all queries. Since novelty is affected by the number of selected passages (smaller sets tend to have higher novelty), we additionally compute size-controlled scores: \textbf{Novel@2} and \textbf{Novel@3} average only over queries where the selected set size is exactly 2 or 3, respectively. 
This allows fair comparison of redundancy removal independent of selection size.

\section{Prompts}
\label{sec:appendix-prompts}
All the prompts we used are presented as follows:

\begin{table*}[h!t]
    \centering
    \begin{tabularx}{0.925\linewidth}{>{\raggedright\arraybackslash}X}
        \toprule
        \textbf{Task}:Prompt used by ``Expanding’’ stage \\
        \midrule
Generate all questions needed to find every piece of information required to answer the final question.

Search Query: $\{question\}$ \\
Answers: $\{answers\}$ \\
Rules: \\
- Each query must stand alone without referencing previous queries or using pronouns like “it,” “they,” or “that.” Always repeat the necessary entities. \\
Output format exactly:\\
"\#\#\# Queries: <Query 1>\textbackslash n<Query 2>... \textbackslash n<Query n>\textbackslash n" \\ 
    \bottomrule
    \end{tabularx}
    \caption{The prompt used by ``Expansion’’ stage.}
    \label{tab:prompt-expand}
\end{table*}

\begin{table*}[h!t]
    \centering
    \begin{tabularx}{0.925\linewidth}{>{\raggedright\arraybackslash}X}
        \toprule
        \textbf{Task}:Prompt used by ``Sampling’’ stage \\
        \midrule
I will provide you with {num} passages, each indicated by a numerical identifier []. \\
Select the passages based on their relevance to the search query: $\{question\}$.\\
$\{context\}$\textbackslash n \\
Search Query: $\{question\}$ \\
Sub-Queries:\textbackslash n$\{queries\}$ \\
Answers: $\{answers\}$ \\
Please follow the steps below:\\
Step 1. Please list up the information requirements to answer the query and sub-queries.\\
Step 2. For each requirement in Step 1, find the passages that has the information of the requirement.\\
Step 3. Choose the passages that mostly covers clear and diverse information to answer the query. Number of passages is unlimited. The format of final output should be '\#\#\# Final Selection: [] [].\textbackslash n', e.g., '\#\#\# Final Selection: [2] [1].\textbackslash n'. \\
    \bottomrule
    \end{tabularx}
    \caption{The prompt used by ``Selection’’ stage.}
    \label{tab:prompt-sampling}
\end{table*}

\begin{table*}[h!t]
    \centering
    \begin{tabularx}{0.925\linewidth}{>{\raggedright\arraybackslash}X}
        \toprule
        \textbf{Task}:Prompt used by ``Refining’’ stage \\
        \midrule
I will provide you with {num} passages, each indicated by a numerical identifier [].\\ 
Select the passages based on their relevance to the search query: $\{question\}$.\\
$\{context\}$\textbackslash n\\
Search Query: $\{question\}$\textbackslash n \\
Answers: $\{answers\}$ \\
Step 1. Identify whether the above set of paragraphs contains irrelevant or repetitive paragraphs.\\
Step 2. Exclude a small amount of irrelevant or repetitive information. The format of final output should be '\#\#\# Final Selection: [] [].\textbackslash n', e.g., '\#\#\# Final Selection: [2] [1].\textbackslash n'. \\ 
    \bottomrule
    \end{tabularx}
    \caption{The prompt used by ``Refinement’’ stage.}
    \label{tab:prompt-refine}
\end{table*}

\begin{table*}[h!t]
    \centering
    \begin{tabularx}{0.925\linewidth}{>{\raggedright\arraybackslash}X}
        \toprule
        \textbf{Task}:Prompt used by ``Answer’’ \\
        \midrule
        $\{context\}$\textbackslash n\\
        Answer the question below concisely in a few words.\\
        Question: $\{question\}$\textbackslash n\\ 
    \bottomrule
    \end{tabularx}
    \caption{The prompt used for ``Question Answering’’.}
    \label{tab:prompt-answer}
\end{table*}

\begin{table*}[h!t]
    \centering
    \begin{tabularx}{0.925\linewidth}{>{\raggedright\arraybackslash}X}
        \toprule
        \textbf{Task}:Prompt used in ``Training’’ \\
        \midrule
I will provide you with $\{num\}$ passages, each indicated by a numerical identifier []. \\
Select the passages based on their relevance to the search query: $\{question\}$.\\
$\{context\}$\textbackslash n\\
Search Query: $\{question\}$\textbackslash n\\ \\
Please follow the steps below:\\
Step 1. Please list up the information requirements to answer the query and sub-queries.\\
Step 2. For each requirement in Step 1, find the passages that has the information of the requirement.\\
Step 3. Choose the passages that mostly covers clear and diverse information to answer the query. Number of passages is unlimited. The format of final output should be '\#\#\# Final Selection: [] [].\textbackslash n', e.g., '\#\#\# Final Selection: [2] [1].\textbackslash n'. \textbackslash n\\ \\
Output only the required format. No explanations, no headings, no bullets, no markdown. \\ 
    \bottomrule
    \end{tabularx}
    \caption{The prompt used in Selector ``Training’’ and  ``inference’’.}
    \label{tab:prompt-infer}
\end{table*}

\end{document}